# Network Level Spatial Temporal Traffic Forecasting with Hierarchical Attention LSTM (HierAttnLSTM)

Tianya Zhang, Ph.D.
Research Assistant Professor
Center for Urban Informatics and Progress
University of Tennessee at Chattanooga
Chattanooga, TN, US, 37405
Email: tianya-zhang@utc.edu
ORCID: 0000-0002-7606-9886

## Abstract

Traffic state data, such as speed, density, volume and travel time collected from ubiquitous roadway detectors require advanced network level analytics for forecasting and identifying significant traffic patterns. This paper explored diverse traffic state datasets from the Caltrans Performance Measurement System (PeMS) hosted on the open benchmark and achieved promising performance compared to well recognized spatial-temporal prediction models. Drawing inspiration from the success of hierarchical architectures in various Artificial Intelligence (AI) tasks, we integrate cell and hidden states from low-level to high-level Long Short-Term Memory (LSTM) networks with the attention pooling mechanism, similar to human perception systems. The developed hierarchical structure is designed to account for dependencies across different time scales, capturing the spatial-temporal correlations of network-level traffic states, enabling the prediction of traffic states for all corridors rather than a single link or route. The efficiency of designed hierarchical LSTM is analyzed by ablation study, demonstrating that the attention-pooling mechanism in both cell and hidden states not only provides higher prediction accuracy but also effectively forecasts unusual congestion patterns. Data and code are made publicly available to support reproducible scientific research.

## Main Text

### 1. Introduction

Traffic state information is a critical component of advanced traveler information systems (ATIS), which are extensively used for route guidance and mode choice. Short-term traffic state (less than 60-minute time horizon) prediction models are essential for trip planning, as they forecast traffic conditions in the near future to help users avoid unexpected delays. Continuous and updated traffic state data enables mobility management centers and commercial navigation apps to effectively adjust their forecasts of network congestion for travelers. From the users' perspective, predictive traffic information is used to select routes, travel modes, and departure times based on perceived certainty. From the system perspective, predicting traffic states allows mobility system engineers to evaluate the potential benefits of various response strategies under different circumstances.

Traffic conditions are influenced by the imbalance between traffic demand and supply, traffic control measures, accidents, as well as external factors such as weather conditions and emergencies. Traditional time series models rely heavily on preprocessing and feature engineering, which is advantageous when the data volume is small. However, traditional statistical prediction methods, with their limited number of parameters, require frequent retraining and are thus inefficient for application across entire roadway networks. Recurrent neural networks (RNNs) address these limitations with automatic feature extraction capabilities. To mitigate the issues of gradient exploding/vanishing, gated mechanisms have been proposed in popular RNN architectures. The challenge of transportation spatial temporal prediction lies not only in complex temporal dependencies but also in capturing and modeling intricate, nonlocal, and nonlinear spatial dependencies between traffic conditions at various locations. Graph Neural Networks (GNNs) tackle traffic network-level prediction challenges in capturing and modeling complex spatial dependencies that traditional methods struggle with. However, their ability to learn dynamic graphs relies on feature engineering to build different node/edge attributes for making informed predictions.

Existing traffic prediction models using LSTM as the backbone typically employ a stacked architecture without cell-state and hidden-state hierarchical feature extraction capabilities. We proposed a hierarchical pooling module to capture information from different time steps, akin to the human perception system that consolidates low-level inputs into high-level abstractions, enhancing robustness and accuracy. The HierAttnLSTM model introduces a novel hierarchical pooling module that distinctly processes hidden states and cell states across LSTM layers, enhancing the capture of complex temporal patterns. Hidden states from the lower layer are pooled to form new input sequences for the upper layer, enabling multi-scale temporal processing and the creation of higher-level abstractions. Uniquely, the model also pools cell states from both lower and upper layers, facilitating the integration of long-term dependencies and ensuring crucial information is preserved across the hierarchy, resulting in a more robust and effective modeling of intricate temporal relationships. This dual pooling mechanism creates a more robust connection between the layers, allowing for better information flow and more effective learning of complex temporal dependencies.

The motivation behind developing a hierarchical LSTM model for traffic state prediction stems from the observation that intelligent perceptual tasks, such as vision and language modeling, benefit from hierarchical representations. Features in successive stages become increasingly global, invariant, and abstract. This theoretical and empirical evidence suggests that a multi-stage hierarchy of representations can improve performance in understanding complex patterns and making accurate predictions. The hierarchical attention-pooling-based LSTM model is designed to learn representations at multiple levels of abstraction. Lower levels of the hierarchy capture local features and dependencies over short time intervals, while higher levels capture more global and long-term patterns. This design enables the model to understand complex temporal relationships, recurring traffic patterns, and other factors affecting traffic states.

## 2. Literature Review

**a.** Traffic Forecasting Models

The field of traffic forecasting has evolved significantly over the years, reflecting advancements in data analysis and computational techniques. Initially, traditional statistical methods were employed to predict traffic patterns, which were built on hand-engineered task-specific parameters include linear regression methods [1, 2], ARIMA [3,4], Kalman filter [5-7], Hidden Markov Models (HMMs) [8], and dynamic Bayesian networks [9]. As technology progressed, machine learning algorithms gained prominence, offering improved accuracy and the ability to handle more complex data. Machine Learning methods include Random Forests [10], support vector regression [11,12], k-Nearest Neighbor (KNN) Methods [13]. The congestion map-based method [14] combines historical data with real-time data to predict travel time. The historical data were classified with Gaussian Mixture Model and K-means algorithm to estimate congestion propagation using a consensual day. Dynamic linear models (DLMs) were designed [15] to approximate the non-linear traffic states. The DLMs assume their model parameters are constantly changing over time, which is used to describe the Spatial-temporal characteristics of temporal traffic data.

The advent of deep learning marked a significant milestone, with Convolutional Neural Networks (CNNs) and Recurrent Neural Networks (RNNs) demonstrating remarkable capabilities in capturing spatial and temporal dependencies in traffic data. Those data-driven approaches don't require location-specific info or strong modeling assumptions, which can fit into the constantly evolving temporal data analysis techniques. A Stacked Auto Encoder (SAE) [16] deep learning method for traffic flow prediction that leverages stacked autoencoders to learn generic traffic flow features, demonstrating superior performance compared to traditional methods. The CRS-ConvLSTM model [17] enhances short-term traffic prediction by identifying critical road sections through a spatiotemporal correlation algorithm and using their traffic speeds as input to a ConvLSTM network. The DMVST-Net [18] enhances taxi demand prediction by integrating temporal (LSTM), spatial (local CNN), and semantic views to capture complex non-linear spatial-temporal relationships in large-scale taxi demand data, outperforming existing methods that consider spatial and temporal aspects independently. The Sequence-to-sequence (Seq2Seq) RNN-based approaches can go beyond the univariate forecasting that outputs network scale travel time prediction. A stacked bidirectional LSTM [19] for network level traffic forecasting that handles missing values with imputation units. Given the success of the Attention mechanism in many fields, this study [20] integrated the attention mechanism with the LSTM model to construct the depth of LSTM and model the long-range dependence.

More recently, Spatial-Temporal Graph Neural Networks (GNNs) have emerged as powerful tools for modeling the inherent network structure of transportation systems, while attention-based architectures have shown promise in focusing on the most relevant features for prediction. DCRNN [21] models traffic flow as a diffusion process on a directed graph, capturing spatial dependencies through bidirectional random walks and temporal dependencies using an encoder-decoder architecture. The Graph WaveNet [22] architecture addresses limitations in spatial-temporal graph modeling by employing an adaptive dependency matrix to capture hidden spatial relationships and utilizing stacked dilated 1D convolutions to handle long temporal sequences. The ASTGCN [23] model enhances traffic flow forecasting by incorporating three independent components to capture recent, daily periodic, and weekly-periodic dependencies with spatial-temporal attention mechanisms. The GCGA [24] addresses the real-time traffic speed estimation problem with limited data, leveraging graph convolution and generative adversarial networks to effectively extract spatial features and generate accurate traffic speed maps. STSGCN model [25] improves spatial-temporal network data forecasting by simultaneously capturing complex localized spatial-temporal correlations and heterogeneities through a synchronous modeling mechanism and multiple time-period modules. The LSGCN [26] framework enhances both long-term and short-term traffic prediction by integrating a novel cosAtt graph attention network with graph convolution networks in a spatial gated block, combined with gated linear units convolution. The GMAN [27] enhances long-term traffic prediction by utilizing an encoder-decoder architecture with multiple spatio-temporal attention blocks and a transform attention layer. SimST [28] replaces computationally expensive Graph Neural Networks (GNNs) with efficient spatial context injectors. This STPGNN [29] introduces a pivotal node identification module, a pivotal graph convolution module, and a parallel framework to effectively capture spatio-temporal traffic features on both pivotal and non-pivotal nodes.

Researchers have also explored hybrid approaches (e.g., DNN-BTF [30], ST-GAT [31], Frigate [32]), combining different methodologies to leverage their respective strengths and address the multifaceted nature of traffic dynamics. Transformer-based models [33-35], Reinforcement Learning [36], ODE-based [37, 38] and Generative Adversarial Networks (GANs) [39] were also applied to spatial-temporal traffic forecasting tasks. This ongoing evolution reflects the continuous effort to improve the accuracy and reliability of traffic state forecasting models, crucial for effective traffic management and urban planning.

### b. Hierarchical Spatial-Temporal Modeling

Hierarchical deep learning architecture is a widely adopted framework for spatial-temporal data analysis, which has been applied in many vision and language learning tasks [40]. Inspired by the success of pyramid feature extraction in computer vision, researchers have tried similar approaches for time series data modeling, and many results have shown great benefits by employing multiscale scheme for efficient video-summarization applications. By incorporating temporal structure with deep ConvNets for video representation for video content analysis, Hierarchical Recurrent Neural Encoder (HRNE) [41] is proposed that can efficiently exploit video temporal structure to model the temporal transitions between frames as well as the transitions between segments. The Temporal Shift Module (TSM) was proposed [42] for hardware-efficient video streaming understanding. TSM model has three main advantages: low latency inference, low memory consumption, and multi-level fusion. A spatial-temporal action detection and localization model [43] using a Hierarchical LSTM and achieved the state-of-the-art in spatial-temporal video analysis, which is a basic functional block for a holistic video understanding and human-machine interaction. The multi-resolution convolutional autoencoder (MrCAE) architecture [44] models the Spatial-temporal dynamics using a progressive-refinement strategy. A multiscale convolutional LSTM network (MultiConvLSTM) [45] was implemented for travel demand and Origin-Destination predictions. Their experiments on real-world New York taxi data have shown that the MultiConvLSTM considers both temporal and spatial correlations and outperforms the existing methods. A deep hierarchical LSTM network [46] for video summarization (DHAVS) extracts spatial-temporal features and applied an attention-based hierarchical LSTM module to capture the temporal dependencies among video frames. Hierarchical spatial-temporal modeling was explored in smart manufacturing in characterizing and monitoring global anomalies to improve higher product quality [47]. The Hierarchical Information Enhanced Spatio-Temporal (HIEST) [48] prediction method improves traffic forecasting by modeling sensor dependencies at regional and global levels, using Meta GCN for node calibration and cross-hierarchy graph convolution for information propagation.

Vision and language understanding task is deemed as the benchmark for evaluating progress in artificial intelligence. Given the impressive performance of hierarchical feature learning in various vision-language understanding applications, in the next section, we propose a novel hierarchical LSTM model for the short-term travel time prediction task. Compared to existing LSTM-based models that only modify the data input layers for feature extraction; our new designed hierarchical LSTM model breaks the interconnections within the "black-box" neural networks. In contrast to GNN models, our method is a plug-and-play solution that requires no feature engineering efforts.

## 3. Methods

### a. Preliminary

In this section, we briefly describe the key components and variants of the building LSTM unit. LSTM is modified from the vanilla RNN (Recurrent Neural Network) model to enhance the capability of long-term temporal dependence for sequential feature extraction. LSTM has shown great performance on many language tasks or time-varying data modeling. The classic LSTM cell has led to several variants by adding new modifications, such as ConvLSTM [49], Grid LSTM [50], and Eidetic LSTM [51]. Three main gates were collectively used for progressively updating the output: Input Gate, Output Gate, and Forget Gate. The key feature of LSTM is the Cell State, which works as the memory pipe to transmit the long-term memory stored in the previous state to the current state. The input and forget gates are used as knobs to determine which information needs to be deleted or added to the cell state. Equation (5) describes how the current cell state adds or forgets information with the forget gate and the input gates. The output gate takes the inputs, newly updated long-term memory, and previous short-term memory to compute a new hidden state/short-term memory. The LSTM unit model (Figure 1) used in this paper is iterated as follows:

$$\mathbf{i}_t = \sigma(\mathrm{W}_{ix}\mathbf{x}_t + \mathbf{b}_{ii} + \mathrm{W}_{ih}\mathbf{h}_{t-1} + \mathbf{b}_{hi}) \tag{1}$$

$$\mathbf{f}_t = \sigma(\mathrm{W}_{fx}\mathbf{x}_t + \mathbf{b}_{if} + \mathrm{W}_{fh}\mathbf{h}_{t-1} + \mathbf{b}_{hf}) \tag{2}$$

$$\mathbf{o}_\mathrm{t} = \sigma(\mathrm{W}_{ox}\mathbf{x}_t + \mathbf{b}_{io} + \mathrm{W}_{oh}\mathbf{h}_{t-1} + \mathbf{b}_{ho}) \tag{3}$$

$$\mathbf{g}_\mathbf{t} = \emptyset(\mathrm{W}_{gx}\mathbf{x}_t + \mathbf{b}_{ig} + \mathrm{W}_{gh}\mathbf{h}_{t-1} + \mathbf{b}_{hg}) \tag{4}$$

$$\mathbf{c}_\mathrm{t} = \mathbf{f}_t \odot \mathbf{c}_{t-1} + \mathbf{i}_t \odot \mathbf{g}_\mathbf{t} \tag{5}$$

$$\mathbf{h}_t = \mathbf{o}_t \odot \emptyset(\mathbf{c}_t) \tag{6}$$

where $f_t$ is forget gate at timestamp $t$, $c_t$ is the cell state at timestamp $t$, and $o_t$ is the output gate at timestamp $t$. $\sigma$ represents a sigmoid operation and $\emptyset$ represents $tanh$ activation function. $\odot$ is the Hadamard product. $W$ is the weight matrix that conducts affine transformation on the input $\boldsymbol{x}_t$ and hidden state $\boldsymbol{h}_t$. Matrices are depicted with capital letters while vectors with non-capital bold letters.

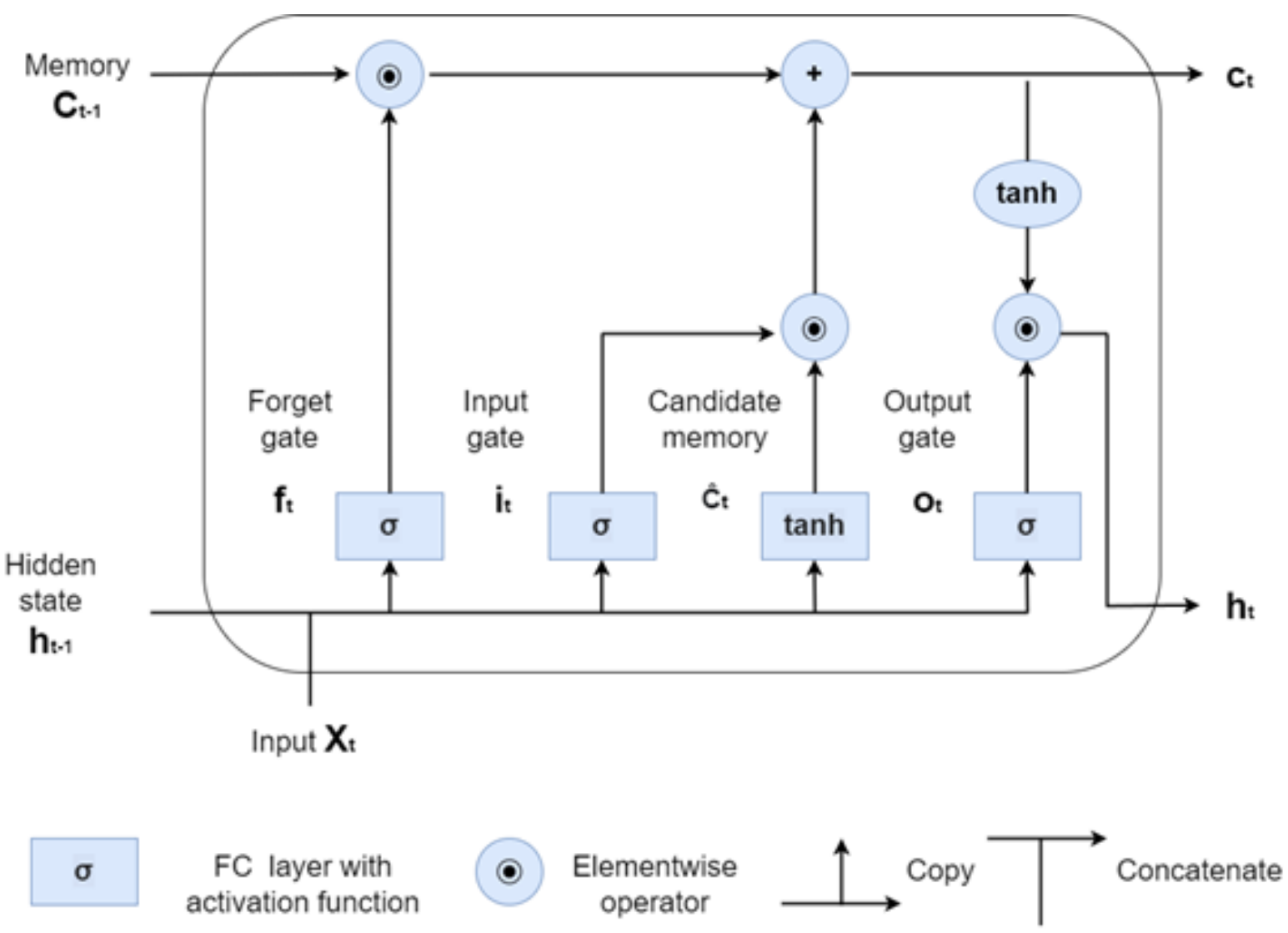

Figure 1 LSTM Unit

## b. Cell and Hidden States Attention Pooling

Figure 2 illustrates our dual attention pooling module, a key innovation in our hierarchical LSTM architecture. This module computes a spatial-temporal representation of network travel time over an extended duration, analogous to pooling operations in CNNs. The superscript $l$ and $l-1$ represent layer number. The subscript $i, i+1, \ldots, i+k$ represent sequential inputs. n-1 and n are timesteps for the upper-layer LSTM. The process begins with input states from the lower layer ($S_i^{l-1}$ *to* $S_{\mathrm{i+k}}^{l-1}$) and, uniquely for cell states, the upper layer ($CS_n^l$, shown by the dashed red line). Each input undergoes an affine transformation, followed by a softmax operation to generate adaptive weights. These weights are then used to create a weighted combination of the original states, producing the final Pooled State ($PS_n^l$). This mechanism effectively increases the temporal receptive field by creating a compact representation of batch sequential inputs.

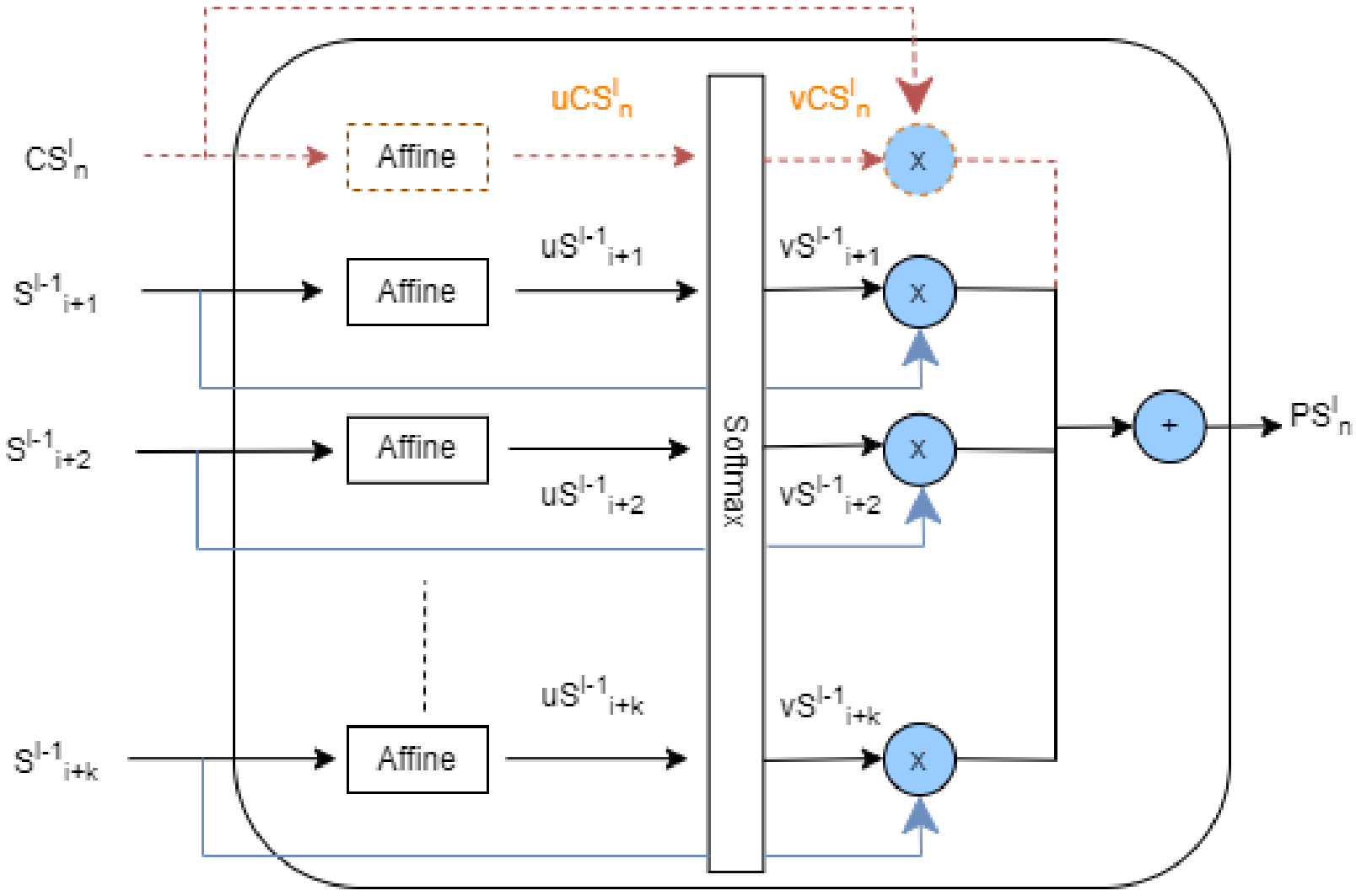

Figure 2. Attention pooling module for hierarchical LSTM. Note the distinct treatment of cell states (dashed red path) incorporating both layers, versus hidden states using only the lower layer.

While hidden states (solid paths) are processed using only information from the lower layer, cell states (dashed red path) uniquely incorporate information from both the current and previous layers. This dual approach allows our model to balance the preservation of long-term dependencies with the creation of hierarchical temporal abstractions. By processing a window of $k$ time steps from the lower layer, the module achieves temporal aggregation, enabling the upper layer to capture more complex temporal patterns. The adaptive weighting through softmax allows the model to focus on the most relevant information across different time steps and states. This sophisticated mechanism enables our hierarchical LSTM to effectively model complex temporal dependencies at various scales, making it particularly suited for predicting network travel times where patterns may exist at both fine-grained and coarse-grained temporal resolutions.

Here's a more detailed explanation of its implementation and integration:

a) **Affine Transformation**: Each input state (cell and hidden) is transformed using an affine transformation to compresses the information in each state vector into a scalar value. With an affine transformation, each cell state ($CS$) and hidden state ($HS$) vector will be converted into a single number:

$$uHS_{i+t}^{l-1} = Affine\left(HS_{i+t}^{l-1}\right) \quad (7)$$

$$uCS_{n-1}^{l} = Affine\left(CS_{n-1}^{l}\right) \quad (8)$$

$$uCS_{i+t}^{l-1} = Affine\left(CS_{i+t}^{l-1}\right) \quad (9)$$

Where:

the prefix $u$ denotes a natural number obtained through Affine Transformation. $CS_{n-1}^{l}$ refers to the cell state of the previous time step in the current layer $l$

$CS_{i+t}^{l-1}$ refers to cell state of lower layer $l-1$ at timestep $i+t$

$HS_{i+t}^{l-1}$ refers to a hidden state in the lower layer $l-1$, at timestep $i+t$

b) **Softmax Weighting**: The transformed values are then passed through a softmax function to compute weight factors. This step determines the relative importance of each state in the pooling process. After affine transformation, the $uCS_n^l$, $uHS_{i+t}^{l-1}$, and $uCS_{i+t}^{l-1}$ will be sent to Softmax to compute weight factors.

$$vHS_{i+t}^{l-1} = \frac{\exp\left(uHS_{i+t}^{l-1}\right)}{\exp\left(\sum_{i=1}^{t=K} uHS_{i+t}^{l-1}\right)} \quad (10)$$

$$vCS_{i+t}^{l-1} = \frac{\exp\left(vCS_{i+t}^{l-1}\right)}{\exp\left(uCS_{n-1}^{l}\right)+\sum_{t=1}^{t=K} \exp\left(uCS_{i+t}^{l-1}\right)} \quad (11)$$

$$vCS_{n-1}^{l} = \frac{\exp\left(uCS_{n-1}^{l}\right)}{\exp\left(uCS_{n-1}^{l}\right)+\sum_{i=1}^{t=K} \exp\left(uCS_{i+t}^{l-1}\right)} \quad (12)$$

The prefix $v$ denotes the weight factor for corresponding Cell States and Hidden States after Softmax operation. The original state vectors will be multiplied by their corresponding weight factors and summed to produce the pooled cell state (PCS) and pooled hidden state (PHS) in the next subsections.

c) **Hidden State Attention Pooling**

The hidden states from the lower layer LSTM are processed through an attention mechanism to create a pooled representation:

$$PHS_n^l = \sum_{t=1}^{t=K} vHS_{i+t}^{l-1} * HS_{i+t}^{l-1} \quad (13)$$

Where:

$PHS_n^l$ is the pooled hidden state (PHS) for the nth time step of the $l^{th}$ layer

$vHS_{i+t}^{l-1}$ is the attention weight for the hidden state at time step $i+t$ of the $(l-1)^{th}$ layer

$HS_{i+t}^{l-1}$ is the hidden state at time step i+t of the $(l-1)^{th}$ layer

$K$ is the number of time steps considered in the pooling window

This pooled hidden state serves as the input to the upper layer LSTM, allowing it to process a more compact and informative representation of the lower layer's output.

d) **Cell State Attention Pooling**

The cell states from both layers are pooled using a similar attention mechanism, but with a critical difference - the inclusion of the upper layer's previous cell state:

$$PCS_n^l = vCS_{n-1}^l * CS_{n-1}^l + \sum_{t=1}^{t=K} vCS_{i+t}^{l-1} * CS_{i+t}^{l-1} \quad (14)$$

Where:

$PCS_n^l$ is the pooled cell state for the $n^{th}$ time step of the $l^{th}$ layer

$vCS_{n-1}^l$ is the attention weight for the previous cell state of the $l^{th}$ layer

$CS_{n-1}^{l}$ is the previous cell state of the $l^{th}$ layer

$vCS_{i+t}^{l-1}$ is the attention weight for the cell state at time step $i + t$ of the $(l - 1)^{th}$ layer

$CS_{i+t}^{l-1}$ is the cell state at time step $i + t$ of the $(l - 1)^{th}$ layer

This unique pooling of cell states allows the model to maintain long-term dependencies from the lower layer. Integrating this information with the existing long-term memory of the upper layer will create a more comprehensive representation of the overall temporal context.

e) **Upper Layer LSTM Update:** The top layer LSTM is updated with $PCS$ and $PHS$ by following equations:

$$\mathbf{i}_n^l = \sigma(\mathrm{W}_{ix}^l PHS_n^l + \mathbf{b}_{ii}^l + \mathrm{W}_{ih}^l \mathbf{h}_{n-1}^l + \mathbf{b}_{hi}^l) \quad (15)$$

$$\mathbf{f}_n^l = \sigma(\mathrm{W}_{fx}^l PHS_n^l + \mathbf{b}_{if}^l + \mathrm{W}_{fh}^l \mathbf{h}_{n-1}^l + \mathbf{b}_{hf}^l) \quad (16)$$

$$\mathbf{o}_\mathrm{n}^l = \sigma(\mathrm{W}_{ox}^l PHS_n^l + \mathbf{b}_{io}^l + \mathrm{W}_{oh}^l \mathbf{h}_{n-1}^l + \mathbf{b}_{ho}^l) \quad (17)$$

$$\mathbf{g}_\mathbf{n}^{\boldsymbol{l}} = \emptyset(\mathrm{W}_{gx}^l PHS_n^l + \mathbf{b}_{ig}^l + \mathrm{W}_{gh}^l \mathbf{h}_{n-1}^l + \mathbf{b}_{hg}^l) \quad (18)$$

$$\mathbf{c}_\mathrm{n}^l = \mathbf{f}_n^l \odot PCS_n^l + \mathbf{i}_n^l \odot \mathbf{g}_\mathbf{n}^{\boldsymbol{l}} \quad (19)$$

$$\mathbf{h}_n^l = \mathbf{o}_t^l \odot \emptyset(\mathbf{c}_n^l) \quad (20)$$

### c. Integration with Multi-Layer LSTM

The integrated structure of the proposed HierAttnLSTM model is shown in Figure 3. In our implementation, we process all spatial nodes simultaneously as the initial input and generate predictions for all nodes across the prediction time window. For hierarchical structure, we employ a dynamic grouping mechanism that adapts to the input sequence length. The stride for grouping lower layers to higher layers is calculated using a custom function that ensures an appropriate balance between detail preservation and computational efficiency.

The lower layer LSTM forms the foundation, processing the entire input sequence and generating hidden states and cell states for each time step. Building upon this, the attention pooling mechanism comes into play, applying its innovative approach to a window of $K$ time steps from the lower layer, while also incorporating the previous cell state from the upper layer. This crucial step leads to the upper layer LSTM, which utilizes the pooled states (PCS and PHS) as its inputs. By doing so, the upper layer effectively processes a more compact, yet information-rich representation of the lower layer's output. This hierarchical structure facilitates temporal abstraction, allowing the upper layer to capture longer-term dependencies and abstract temporal patterns that might be less apparent in the lower layer's more granular output.

The bottom layer processes the entire time sequence, while the upper layer computes the pooled hidden states with latent variables. The cell states that represent the longer memories from both bottom and top layer LSTMs are aggregated by the attention pooling to generate new cell states for top-layer LSTM. Following the hierarchical pooled LSTMs, the model incorporates a self-attention layer for dimension reduction, further distilling the hierarchically processed features. The output from this attention layer is then fed into a fully connected layer for final prediction. This structure allows the hidden states from the lower-level LSTM to serve as new time sequences for the top-level LSTM, while the cell states of the top-layer LSTM are computed as a function of both the previous layer and its own states. Through this sophisticated interplay of layers, attention mechanisms, and state handling, the HierAttnLSTM model achieves a multi-scale approach to spatial and temporal information processing, adeptly capturing complex patterns at various levels of abstraction.

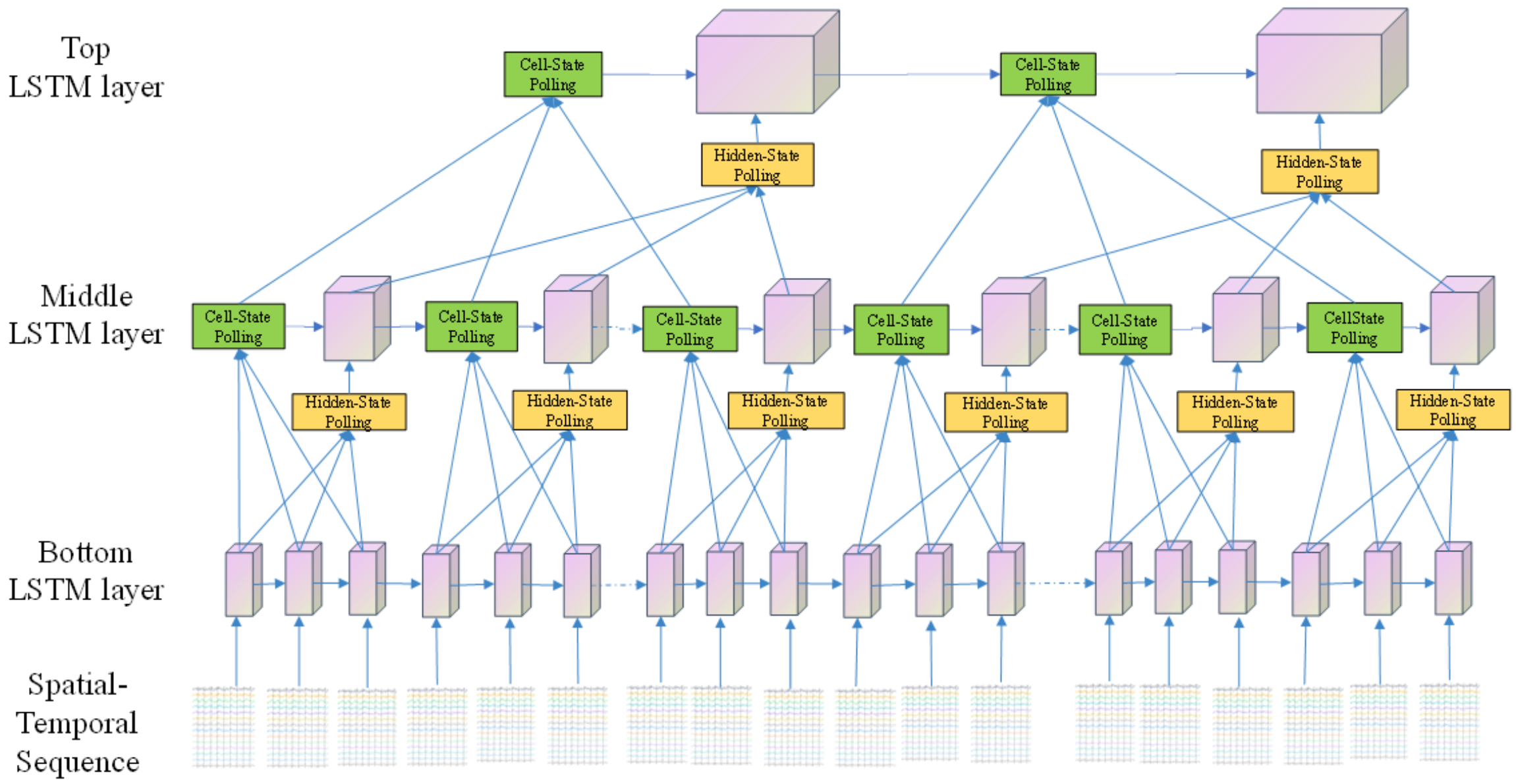

Figure 3 Hierarchical Attention LSTM (HierAttnLSTM)

## 4. Experiment

### a. Dataset

The dataset is collected from Caltrans Performance Measurement System (PeMS), an online system that continuously gathers real-world sensor data, offering a comprehensive and up-to-date representation of traffic conditions [52]. PEMS' public accessibility and widespread use in similar traffic systems allows for easy generalization of model results, increasing the practical impact to a broader range of real-world applications. We used the PeMS-Bay, PeMSD4, and PeMSD8 datasets standardized by the LibCity [53] benchmark. LibCity aims to provide researchers with a reliable experimental tool and a convenient development framework, ensuring standardization and reproducibility in the field of traffic forecasting. In this study, we adopt the structure of LibCity atomic files and apply normalization as the primary preprocessing step, without performing any filtering or data imputation. The datasets PEMSD4, PEMSD8, and PEMS-BAY provide diverse traffic data for prediction tasks. PEMSD4 covers 307 nodes over 16,992 timesteps from January to February 2018, with flow, speed, and occupancy data. PEMSD8 includes 170 nodes over 17,856 timesteps from July to August 2016, also with flow, speed, and occupancy data. PEMS-BAY is the largest, with 325 nodes and 52,116 timesteps from January to May 2017, focusing solely on speed data. While these publicly available datasets cover flow and speed, they lack travel time data, which was addressed by downloading an additional PEMS-Bay dataset for travel time prediction testing from January 2020 to October 2021.

### b. Data Exploratory Analysis

Traffic sensor data provides a comprehensive view of variations in monitored area, revealing clear patterns based on time of day, day of the week, and month of the year. These insights can inform traffic management strategies and help individuals plan their travel more efficiently. The PEMS-BAY area travel time data from 2020 reveals interesting patterns across different time scales. Monthly averages show relatively consistent median travel times throughout the year, with slightly more variability in the early months and lower times in the middle of the year. The monthly data suggests some seasonal effects, with winter months showing more variability. This could be due to weather conditions or holiday-related traffic patterns. Daily patterns demonstrate clear rush hour peaks on weekdays, with Friday evenings experiencing the highest travel times. Weekends exhibit a distinct pattern with less pronounced morning peaks and generally lower travel times compared to weekdays. (see Figure 4)

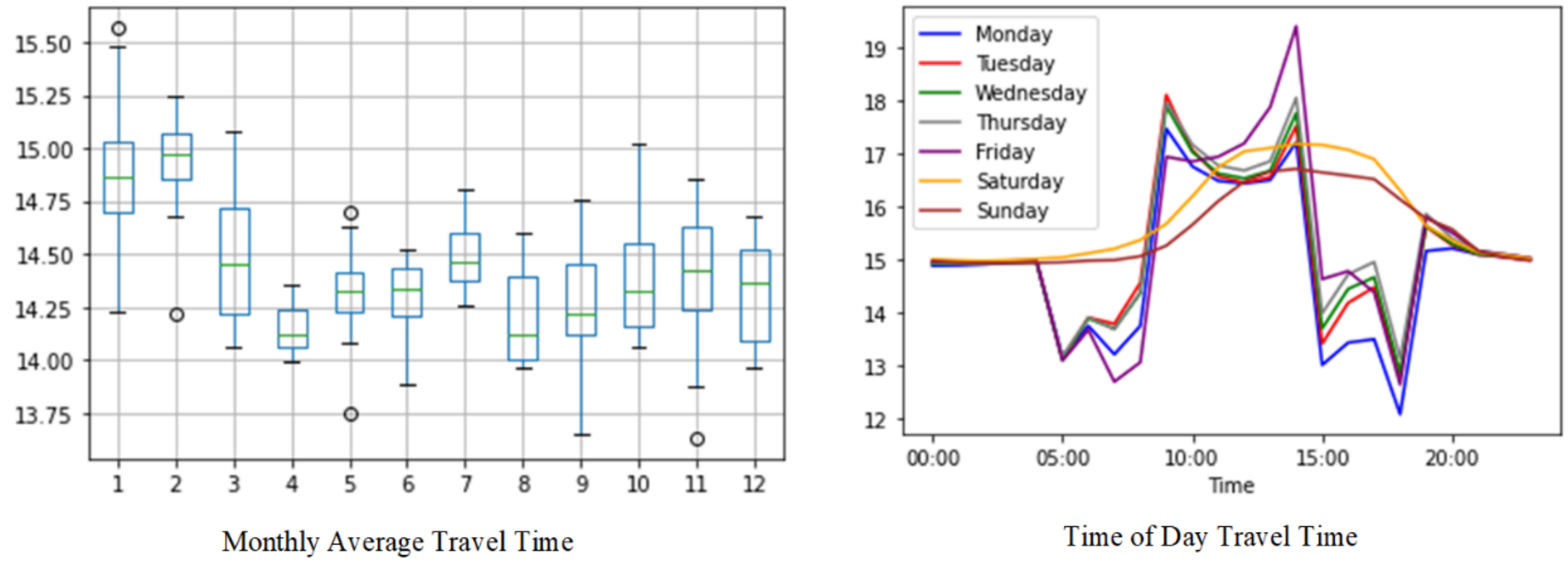

Monthly Average Travel Time

Time of Day Travel Time

Figure 4 Monthly and Daily Travel Time Pattern in PEMS District 4 data

The main feature of time series is autocorrelation (AC), which is the correlation for the data with itself at previous timestamps. It is the assumption of time series forecasting models and helps us reveal the underlying patterns. The partial autocorrelation function (PACF) is similar to the ACF except that it displays only the correlation between two observations. Additionally, analyzing the ACF and PACF in conjunction is necessary for choosing the appropriate model for our time series prediction. A very high autocorrelation in travel time data has been identified after calculating autocorrelation and partial autocorrelation because traffic conditions 5 minutes ago will most likely affect the current travel time. As time increases, the correlation declines more and more (see **Figure 5**).

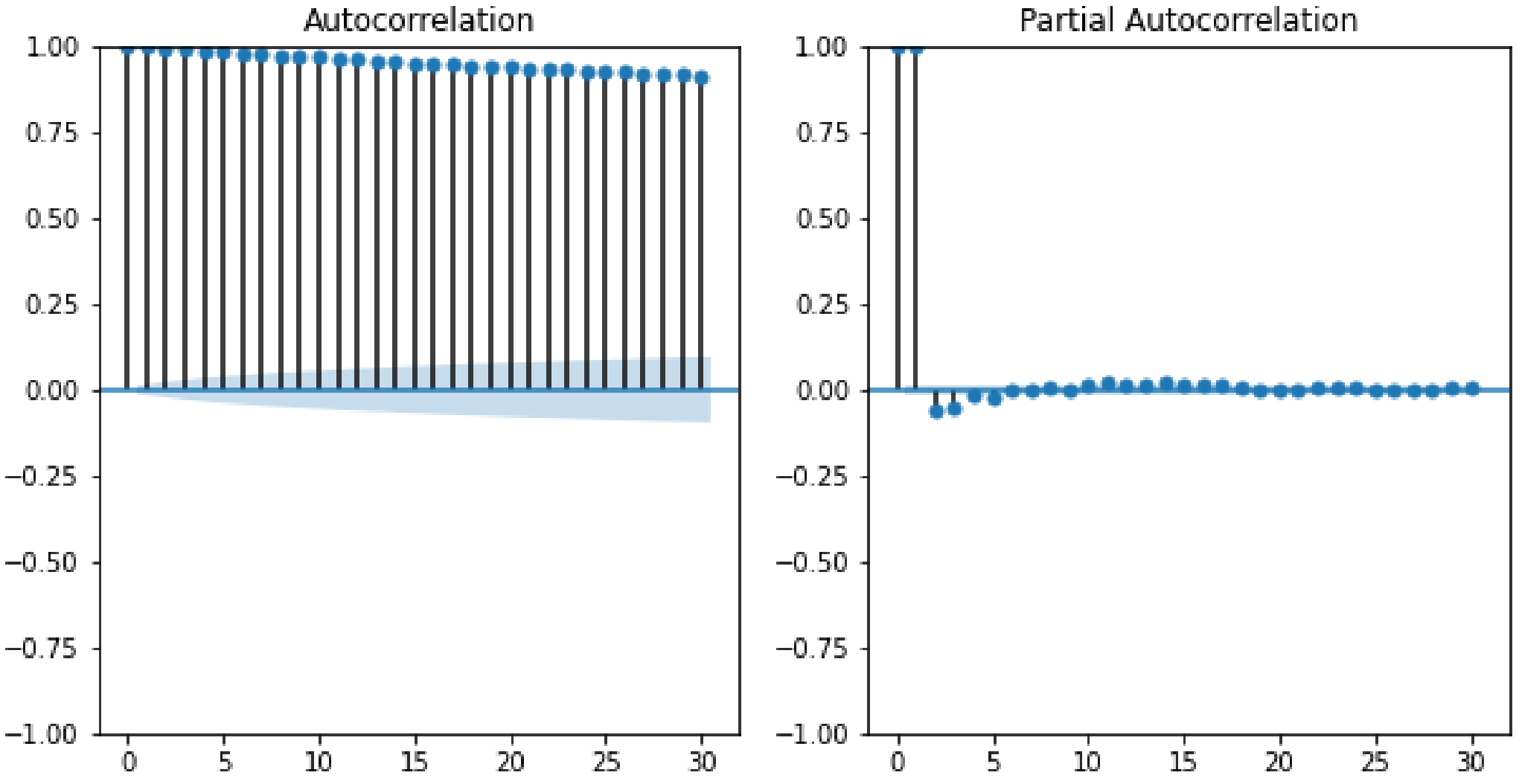

Figure 5 Travel State Data Autocorrelation Analysis

### c. Implementation Details

A comprehensive approach is applied for hyperparameter tuning, systematically exploring different combinations of hidden sizes (64, 128, 256), number of layers (2, 3, 4), and attention hops (2, 3, 4), among other parameters. For each combination, the implementation creates a configuration dictionary with specific model parameters. The training process is managed through an executor configuration file, which specifies key training parameters. The designed model is trained on a Google Cloud A100 GPU, a maximum of 100 epochs with early stopping after 5 epochs of no improvement, and the use of the Adam optimizer with a MultiStepLR learning rate scheduler. The configuration also sets up logging, model saving, and evaluation criteria. This comprehensive setup, implemented using the LibCity library, enables a systematic exploration of the model's hyperparameter space, allowing researchers to identify the most effective configuration for their specific task while maintaining consistency in the training environment and methodology. After completing the hyperparameter tuning process, the best-performing model configuration was identified with a hidden size of 128, 3 layers, and 3 attention hops. This optimal configuration likely provides the best balance of model complexity and performance for the specific task at hand. Two performance metrics are used to evaluate the model's performance. Mean Absolute Error (MAE) is used to measure model accuracy. Root Mean Square Error (RMSE) is sensitive to model stability.

$$RMSE = \sqrt{\frac{1}{N*C}\sum_{j=1}^{C}\sum_{i=1}^{N}\left(\hat{T}_{i,j}(t) - T_{i,j}(t)\right)^{2}} \quad (21)$$

$$MAE = \frac{1}{N*C}\sum_{j=1}^{C}\sum_{i=1}^{N}\left|\hat{T}_{i,j}(t) - T_{i,j}(t)\right| \quad (22)$$

where, $T_{i,j}\ (t)$ and $\hat{T}_{i,j}(t)$ are the predicted and ground truth travel time for corridor $j$ at timestamp $i$. $C$ is the total number of corridors. $N$ is the total number of timestamps in the output window.

## 5. Results

### a. Traffic Flow Prediction

Table 1. PEMSD4 Traffic Flow Forecasting

| MODEL | 3 STEP (15-min) | | 6 STEP (30-min) | | 9 STEP (45-min) | | 12 STEP (60-min) | |
|---|---|---|---|---|---|---|---|---|
| | MAE | RMSE | MAE | RMSE | MAE | RMSE | MAE | RMSE |
| **HierAttLSTM** | **9.079** | **22.766** | **8.933** | **22.574** | **9.076** | **22.884** | **9.168** | **22.844** |
| AGCRN [55] | 18.132 | 29.221 | 18.834 | 30.464 | 19.377 | 31.310 | 19.851 | 31.965 |
| GWNET [22] | 17.692 | 28.516 | 18.574 | 29.888 | 19.247 | 30.895 | 19.956 | 31.848 |
| MTGNN [54] | 17.925 | 28.837 | 18.760 | 30.296 | 19.349 | 31.334 | 20.135 | 32.510 |
| GMAN [27] | 18.790 | 29.549 | 19.538 | 30.805 | 20.189 | 31.765 | 20.865 | 32.575 |
| STGCN [56] | 19.146 | 30.301 | 20.133 | 31.886 | 20.830 | 33.056 | 21.567 | 34.200 |
| GRU [57] | 22.441 | 36.286 | 22.506 | 36.342 | 22.571 | 36.415 | 22.583 | 36.447 |
| Seq2Seq [58] | 22.585 | 36.475 | 22.581 | 36.348 | 22.762 | 36.554 | 23.163 | 36.988 |
| DCRNN [21] | 19.581 | 31.125 | 21.467 | 34.067 | 23.152 | 36.665 | 24.864 | 39.228 |
| STG2Seq [60] | 23.006 | 35.973 | 23.251 | 36.227 | 23.744 | 36.822 | 24.935 | 38.330 |
| AE [59] | 23.999 | 37.942 | 24.024 | 37.990 | 24.401 | 38.446 | 25.025 | 39.289 |
| ASTGCN [23] | 20.530 | 31.755 | 22.971 | 35.033 | 24.982 | 38.170 | 27.495 | 41.776 |
| TGCN [61] | 21.678 | 34.635 | 23.962 | 37.777 | 26.340 | 41.045 | 29.062 | 44.794 |

Table 2. PEMSD8 Traffic Flow Forecasting

| MODEL | 3 STEP (15-min) | | 6 STEP (30-min) | | 9 STEP (45-min) | | 12 STEP (60-min) | |
|---|---|---|---|---|---|---|---|---|
| | MAE | RMSE | MAE | RMSE | MAE | RMSE | MAE | RMSE |
| **HierAttLSTM** | **8.375** | **20.356** | **9.204** | **22.518** | **9.427** | **22.715** | **9.215** | **22.320** |
| GWNET [22] | 13.486 | 21.615 | 14.349 | 23.375 | 15.039 | 24.773 | 15.672 | 25.855 |
| AGCRN [55] | 14.146 | 22.241 | 14.962 | 24.055 | 15.675 | 25.445 | 16.427 | 26.557 |
| MTGNN [54] | 14.001 | 21.988 | 14.883 | 23.624 | 15.707 | 24.873 | 16.583 | 26.128 |
| STGCN [56] | 15.166 | 23.615 | 16.188 | 25.401 | 16.971 | 26.556 | 17.819 | 27.818 |
| GMAN [27] | 15.158 | 23.021 | 15.924 | 24.553 | 16.725 | 25.738 | 17.837 | 27.141 |
| DCRNN [21] | 15.139 | 23.476 | 16.619 | 25.982 | 17.960 | 28.009 | 19.345 | 30.058 |
| Seq2Seq [58] | 19.186 | 31.220 | 19.326 | 31.446 | 19.618 | 31.772 | 19.894 | 32.117 |
| GRU [57] | 19.992 | 32.276 | 20.126 | 32.569 | 20.274 | 32.853 | 20.461 | 33.200 |
| STG2Seq [60] | 18.217 | 27.334 | 19.479 | 29.289 | 20.432 | 30.617 | 21.445 | 32.130 |
| ASTGCN [23] | 16.433 | 24.878 | 18.547 | 27.919 | 20.357 | 30.206 | 22.284 | 32.706 |
| AE [59] | 22.266 | 35.562 | 22.209 | 35.557 | 22.335 | 35.696 | 22.865 | 36.269 |
| TGCN [61] | 17.348 | 25.934 | 19.109 | 28.846 | 21.007 | 31.524 | 23.417 | 34.694 |

The performance of the HierAttnLSTM model on both PEMSD4 (Table 1) and PEMSD8 (Table 2) datasets demonstrates significant improvements over existing baseline models for traffic flow forecasting. Across all forecast horizons (15, 30, 45, and 60 minutes),

HierAttnLSTM consistently outperforms the other 12 models in both Mean Absolute Error (MAE) and Root Mean Square Error (RMSE) metrics. For PEMSD4, the model achieves MAE values ranging from 9.079 to 9.168 and RMSE values from 22.574 to 22.884 across different time steps, substantially outperforming the next best model, AGCRN. Similarly, for PEMSD8, HierAttnLSTM shows remarkable performance with MAE values between 8.375 and 9.427, and RMSE values between 20.356 and 22.715. The improvements are particularly striking for shorter-term predictions, with the 3-step (15-minute) forecast showing nearly 50% reduction in MAE for PEMSD4 and about 38% for PEMSD8 compared to the next best models. These results indicate that HierAttnLSTM achieves state-of-the-art performance in traffic flow forecasting, offering substantial gains in prediction accuracy across different datasets and forecast horizons.

### b. Traffic Speed Prediction

Table 3. PEMS-BAY Traffic Speed Forecasting

| MODEL | 3 STEP (15-min) | | 6 STEP (30-min) | | 9 STEP (45-min) | | 12 STEP (60-min) | |
|---|---|---|---|---|---|---|---|---|
| | MAE | RMSE | MAE | RMSE | MAE | RMSE | MAE | RMSE |
| GWNET [22] | 1.317 | 2.782 | 1.635 | 3.704 | 1.802 | 4.154 | 1.914 | 4.404 |
| MTGNN [54] | 1.331 | 2.797 | 1.657 | 3.760 | 1.831 | 4.214 | 1.954 | 4.489 |
| DCRNN [21] | 1.314 | 2.775 | 1.652 | 3.777 | 1.841 | 4.301 | 1.966 | 4.600 |
| AGCRN [55] | 1.368 | 2.868 | 1.686 | 3.827 | 1.845 | 4.265 | 1.966 | 4.587 |
| STGCN [56] | 1.450 | 2.872 | 1.768 | 3.742 | 1.941 | 4.140 | 2.057 | 4.355 |
| GMAN [27] | 1.521 | 2.950 | 1.828 | 3.733 | 1.998 | 4.107 | 2.115 | 4.321 |
| ASTGCN [23] | 1.497 | 3.024 | 1.954 | 4.091 | 2.253 | 4.708 | 2.522 | 5.172 |
| **HierAttLSTM** | **2.493** | **5.163** | **2.496** | **5.177** | **2.779** | **5.494** | **2.587** | **5.340** |
| GRU [54] | 2.491 | 5.204 | 2.508 | 5.288 | 2.535 | 5.384 | 2.575 | 5.510 |
| Seq2Seq [58] | 2.443 | 5.108 | 2.446 | 5.144 | 2.493 | 5.259 | 2.581 | 5.470 |
| AE [59] | 2.570 | 5.302 | 2.573 | 5.288 | 2.627 | 5.392 | 2.724 | 5.608 |
| STG2Seq [60] | 2.192 | 4.231 | 2.424 | 4.826 | 2.604 | 5.266 | 2.768 | 5.650 |
| TGCN [61] | 2.633 | 5.288 | 2.739 | 5.525 | 2.906 | 5.875 | 3.103 | 6.314 |

The results for PEMS-BAY (Table 3) traffic speed prediction reveal that graph-based models like GWNET, MTGNN, and DCRNN outperform the HierAttnLSTM model across all forecast horizons. The variation in the model's performance across different scenarios can be attributed to preprocessed and normalized. Our experiments revealed that the choice of scaler (e.g., 0-1 normalization, -1 to 1 normalization, or standard normal scaling) can lead to notable performance variations. While HierAttnLSTM shows consistent performance across time steps, it doesn't match the accuracy of several graph-based models on this complex dataset. This outcome highlights a promising future research direction: combining graph models with the Hierarchical Attention LSTM approach. Such a hybrid model could potentially leverage the strengths of both architectures, addressing the current limitations on datasets with complex spatial relationships and improving performance on large-scale traffic networks like PEMS-BAY.

## d. Comparative Analysis

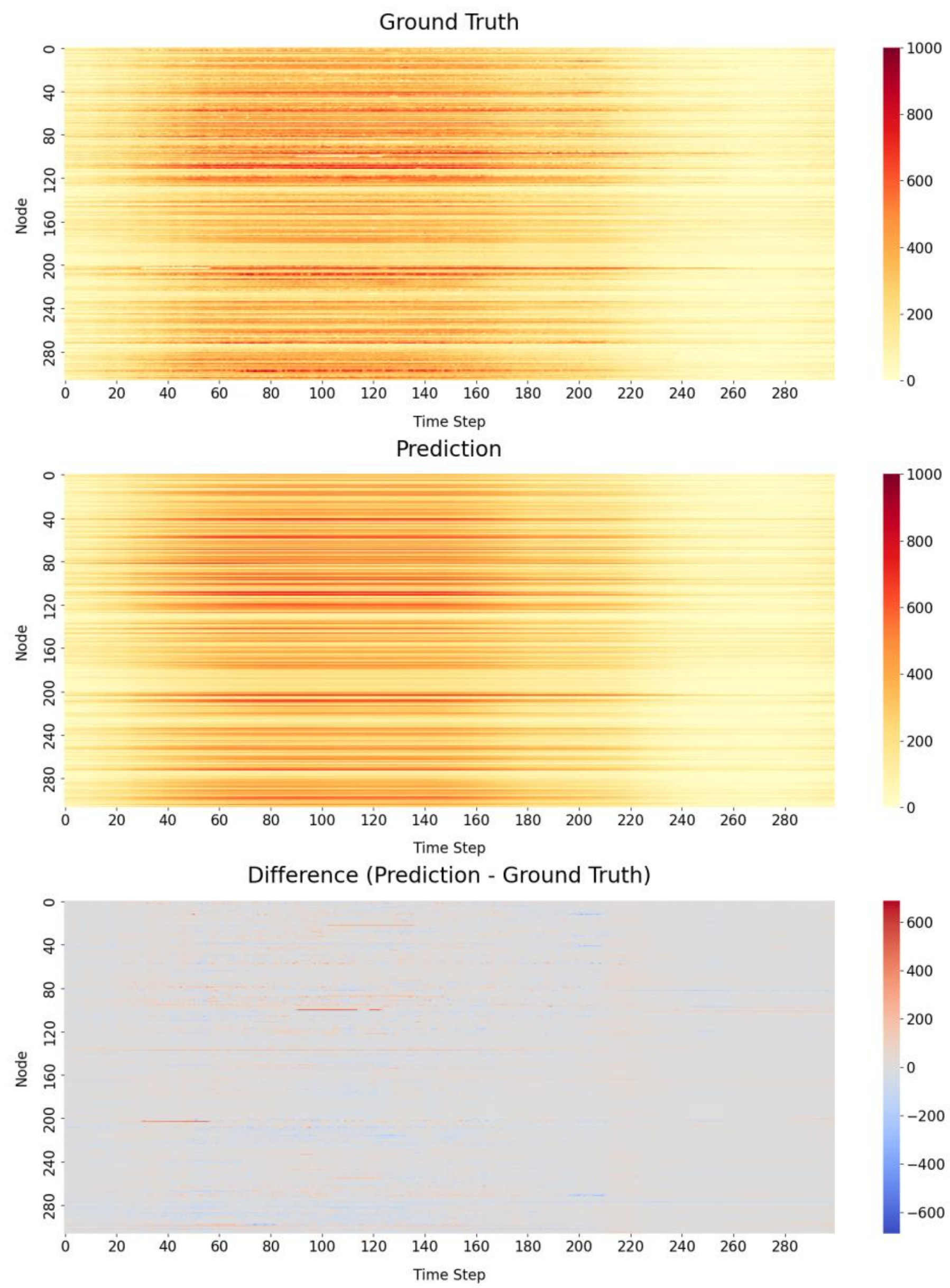

Figure 6. Spatial Temporal Prediction Results Compared to Ground Truth

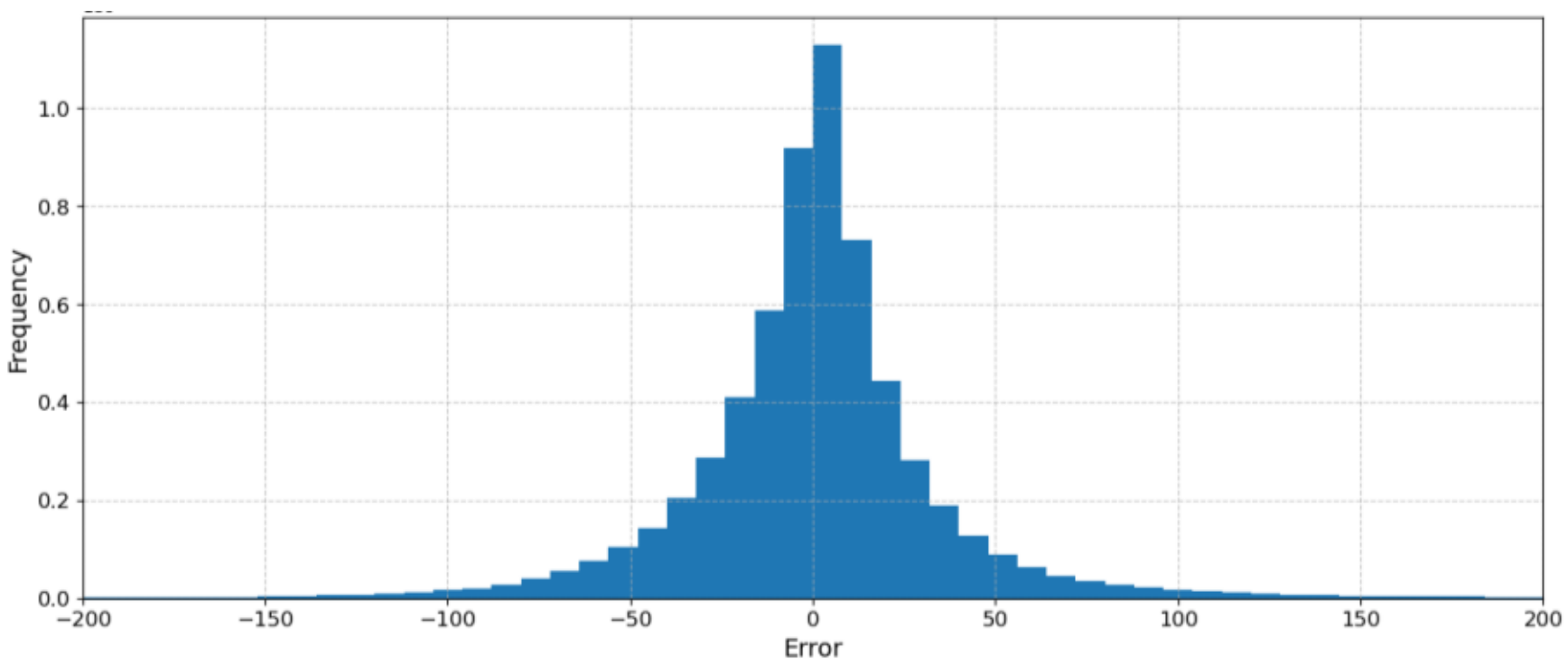

Figure 7. Distribution of Prediction Errors

The prediction results (Figure 6) for the PEMSD4 traffic flow data demonstrate the model's strong performance in capturing both spatial and temporal patterns across 325 nodes and 300 time steps. The error distribution (Figure 7) shows a symmetric, bell-shaped curve centered around zero, indicating unbiased predictions with most errors falling within a small range of -50 to +50 units. This is further supported by the visual similarity between the ground truth and prediction heatmaps, which both display consistent horizontal streaks of higher intensity likely representing busier roads or peak traffic times. The difference heatmap predominantly shows light grey areas, confirming the overall accuracy of the predictions, with only sporadic spots of light red and blue indicating occasional over- or under-predictions. Certain nodes exhibit more variation in prediction accuracy, visible as horizontal streaks in the difference heatmap, suggesting that some locations or road types present greater challenges for the model. Additionally, isolated bright spots in the difference heatmap indicate occasional large errors, though these are rare. The model's performance remains consistent across the entire time range, showing no obvious degradation over time.

Table 4. PEMS-BAY Traffic Speed Forecasting

| Model | Parameter Count | Size (MB) |
|---|---|---|
| MSTGCN | 169596 | 0.65 |
| DCRNN | 372483 | 1.42 |
| GWNET | 410484 | 1.57 |
| HierAttnLstm(64) | 415107 | 1.58 |
| ASTGCN | 556296 | 2.12 |
| AGCRN | 745160 | 2.84 |
| HierAttnLstm(128) | 806917 | 3.08 |
| STGCN | 1476003 | 5.63 |

With a hidden state size of 64, HierAttnLstm(64) maintains a relatively modest model size of 1.58 MB. This places it in the middle range of model complexities, comparable to GWNET but with superior performance. Increasing the hidden state size to 128 results in a larger model (3.08 MB) and yields about 1% improvement. This marginal gain suggests that the smaller version (64 hidden state) might be more cost-effective for most applications. Notably, both versions of HierAttnLstm outperform the other baselines significantly in terms of MSE, despite some models like STGCN having substantially larger parameter counts (1,476,003) and model sizes (5.63 MB), which indicates that HierAttnLstm achieves a better trade-off between model complexity and predictive accuracy.

## 6. Ablation Study

In ablation study, we evaluated the effectiveness of our proposed HierAttnLSTM model against baseline deep learning models for travel time prediction. Additional data extracted from PEMS-BAY were downloaded with our own scraping tool, as public benchmarks lack travel time prediction datasets, we compared our designed LSTM model to vanilla LSTM models (unidirectional Stacked LSTM and Bidirectional Stacked LSTM) implemented without the attention pooling layer for Hidden and Cell States. The baseline models process spatial-temporal input data, with travel time information from all corridors fed at each time step. This comparison aims to isolate the impact of the Attention Pooling Layer in our HierAttnLSTM model, demonstrating its contribution to performance in travel time prediction tasks.

Given all $C$ corridors in the study area and 5-minute resolution data, the previous 2-hour travel time records of corridor$j$ are denoted as $\{tt_j^{T-23}, tt_j^{T-22}, \dots, tt_j^{T}\}$ $(j \in [1, C])$. The deep learning model output is the travel time at future time stamp $T + \delta t$ for all corridors $\{tt_1^{T+\delta t}, tt_2^{T+\delta t}, \dots, tt_C^{T+\delta t}\}$. The training, validation and testing dataset were random generated with sample sizes of 12000 (60%), 4000 (20%) and 4000 (20%). The comparison results are shown in Table 5.

Table 5. Ablation Analysis on Travel Time Prediction at Different Horizons

| Model | 15 min | | 30 min | | 45 min | |
|---|---|---|---|---|---|---|
| | MAE | RMSE | MAE | RMSE | MAE | RMSE |
| Stacked LSTM | 0.247 | 0.445 | 0.272 | 0.517 | 0.286 | 0.557 |
| Stacked BiLSTM | 0.278 | 0.470 | 0.296 | 0.541 | 0.314 | 0.583 |
| **HierAttnLSTM** | **0.195** | **0.339** | **0.235** | **0.424** | **0.268** | **0.49** |

Our model has shown considerably better prediction results than existing LSTM-based travel time prediction results (Table 5). In Figure 8, sample travel time prediction results from different prediction horizons are presented for one-week period. Our proposed model demonstrates significant advantages over two other LSTM-based baselines, after removing the Hierarchical Attention Pooling. More specifically, the HierAttnLSTM can correctly predict the high spikes in travel time in the extended future, which is often the most desirable functionality of travel time prediction model. While the comparable models tend to underestimate the unexpected congestion and fall short of predicting the sudden spikes. The hierarchical attention pooling enhanced the spatial-temporal receptive fields of different levels of LSTM units, which augmented the model's capacity for capturing unusual traffic patterns. The result indicates that adding hierarchical information with attention pooling to distill the cell states of LSTMs could successfully improve the travel time forecasting accuracy.

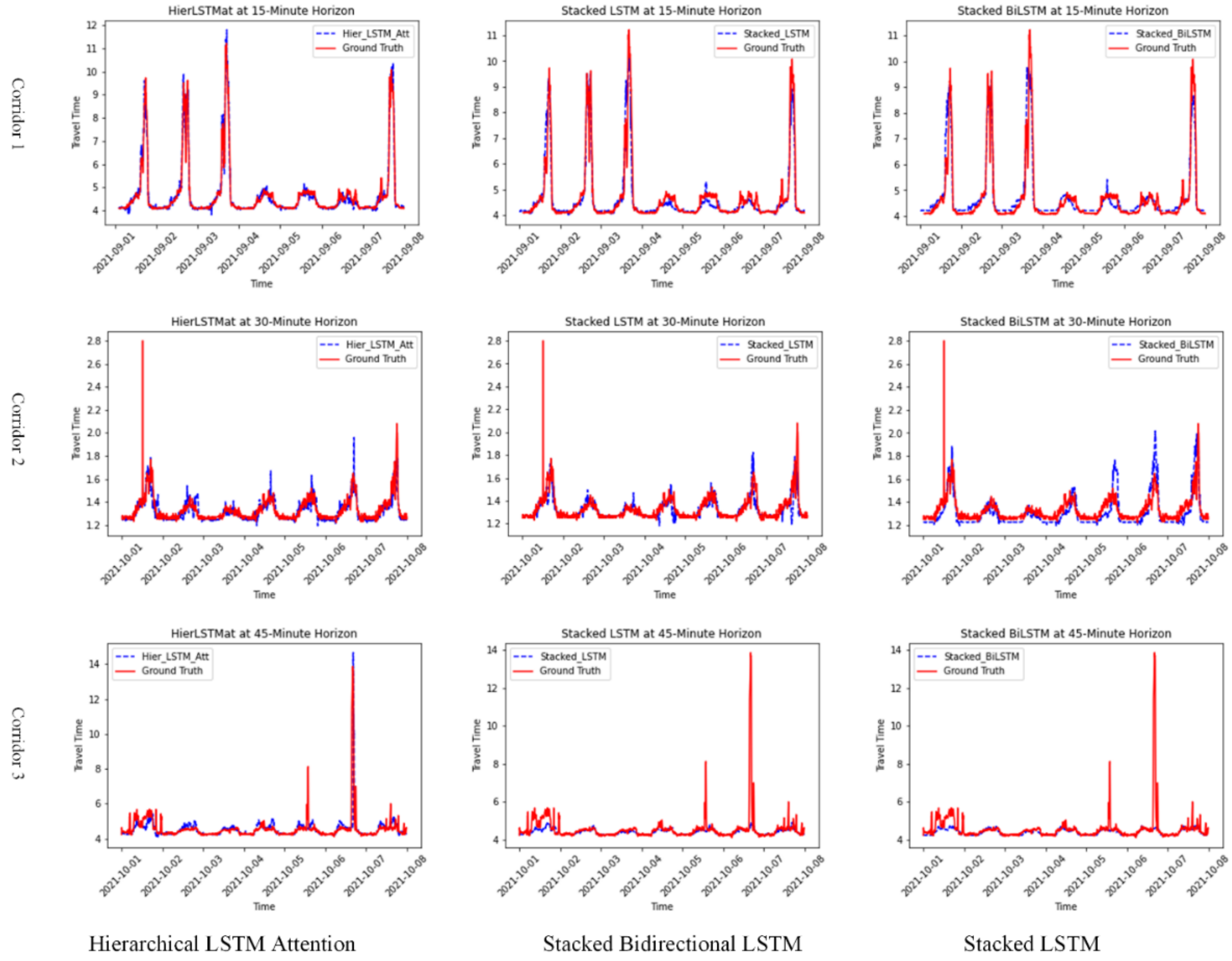

Figure 8. One Week Travel Time Prediction Samples on Self-Downloaded PEMS-BAY Dataset

## 7. Conclusions

The ability to learn hierarchical representations automatically from the data makes the Hierarchical-Attention-LSTM traffic state prediction model a powerful tool for developing accurate and robust travel information prediction systems. From the model design perspective, this paper adds hierarchical feature pooling to the multi-layer LSTM and demonstrates superior prediction accuracy. The proposed Cell and Hidden states pooling architecture ensures that only important features are forwarded from lower to higher layers, mimicking the multiscale information abstraction of the human brain and adaptable to other spatial-temporal learning tasks. Our approach redesigns the internal structure of multi-layer LSTM by introducing attention pooling, which allows the model to better focus on relevant information through novel attention pooling modules. The attention mechanism selectively emphasizes or downplays hidden and cell states based on their importance for predictions, comprehensively leveraging information stored in LSTM cells and improving retention of important contextual information over time.

Importantly, we tested our model on different traffic state prediction tasks: traffic flow, speed, and travel time, using both publicly accessible datasets and our own scraped dataset. This comprehensive evaluation demonstrates the model's versatility and effectiveness across various traffic prediction challenges. Furthermore, we conducted a thorough analysis through ablation studies, clearly demonstrating the effectiveness of adding Dual Pooling to multilayer LSTM. This validation reinforces the key innovation of our approach and its contribution to improved performance. Testing results show the proposed model exhibits the capability to predict unusual spikes in travel time caused by traffic congestion. This crucial finding indicates better generalization to unseen data and more reliable predictions in real-world scenarios.

For future research, exploring additional roadway information with Graph-based methods could further enhance the translation of multi-source data inputs into more abstract representations, potentially leading to even more accurate and robust traffic prediction systems.

## 8. Author contributions

The author confirms sole responsibility for the following: study conception and design, data collection, analysis and interpretation of results, and manuscript preparation.

## 9. Acknowledgments

This research is supported by The Federal Highway Administration (FHWA) Exploratory Advanced Research (EAR) Program. Award No.: 693JJ32350028.

## 10. Conflict of interest

The authors declare that they have no conflict of interest.

## 11. Data availability

The data that support the findings of this study are available in the GitHub repository. These data were derived from the following resources available in the public domain: https://github.com/TeRyZh/Network-Level-Travel-Prediction-Hierarchical-Attention-LSTM.